\documentclass[twoside,11pt]{article}

\usepackage[preprint]{jmlr2e}

\usepackage{subcaption}

\usepackage{amsmath}
\usepackage{amsfonts}
\usepackage{amssymb}

\usepackage{multirow}
\usepackage{makecell}
\usepackage{siunitx}

\usepackage{mathtools}
\mathtoolsset{showonlyrefs,showmanualtags}

\DeclareMathOperator*{\argmax}{\mathrm{arg\,max}}
\newcommand{\Mid}{\;\middle|\;}

\ShortHeadings{Deep Learning for Fast Inference of Mechanistic Models' Parameters}{Borisyak, Born, Neubauer and Cruz-Bournazou}
\firstpageno{1}

\begin{document}

\title{Deep Learning for Fast Inference of Mechanistic Models' Parameters}

\author{\name Maxim Borisyak \email maxim.borisyak@tu-berlin.de\\
		\name Stefan Born \\
		\name Peter Neubauer \\
		\name Mariano Nicolas Cruz-Bournazou\\
		\addr Technische Universität Berlin, Straße 17 des Juni 135, 10623
Berlin, Germany}

\maketitle

\begin{abstract}
Inferring parameters of macro-kinetic growth models, typically
represented by Ordinary Differential Equations (ODE), from the
experimental data is a crucial step in bioprocess engineering.
Conventionally, estimates of the parameters are obtained by fitting the
mechanistic model to observations. Fitting, however, requires a
significant computational power.

Specifically, during the development of new bioprocesses that use
previously unknown organisms or strains, efficient, robust, and
computationally cheap methods for parameter estimation are of great
value. In this work, we propose using Deep Neural Networks (NN) for
directly predicting parameters of mechanistic models given observations.
The approach requires spending computational resources for training a
NN, nonetheless, once trained, such a network can provide parameter
estimates orders of magnitude faster than conventional methods.

We consider a training procedure that combines Neural Networks and
mechanistic models. We demonstrate the performance of the proposed
algorithms on data sampled from several mechanistic models used in
bioengineering describing a typical industrial batch process and compare
the proposed method, a typical gradient-based fitting procedure, and the
combination of the two. We find that, while Neural Network estimates are
slightly improved by further fitting, these estimates are measurably
better than the fitting procedure alone.
\end{abstract}

\section{Introduction}

Mechanistic growth models play an important role in bioprocess
development. These models are derived from the first principles and
their parameters are readily interpretable. Such models also enable
computer-aided design and control of bioprocesses. Most of the models
are expressed as systems of Ordinary Differential Equations (ODE).
Examples include models by~\citet{lin2001determination, neubauer2003metabolic, anane2017modelling}.
Typically, such ODE systems do not admit analytical
solutions and, therefore, one has to rely on dedicated ODE solvers.
Since the systems tend to be highly non-linear, a solver needs to
perform a large number of integration steps. This is especially the
case, when models contain both, fast and slow dynamics~\citep{anane2017modelling}.

Estimation of mechanistic models' parameters is conventionally done by
fitting the model to observations following the Maximum Likelihood
principle:
\begin{align}
	L(\theta) = \log P(x \mid \theta) \to \max; \label{eq:ml}
\end{align}
where $x, \theta$ are observations and parameters of the model.

\begin{figure}{t}
	\centering
	\includegraphics[width=\textwidth]{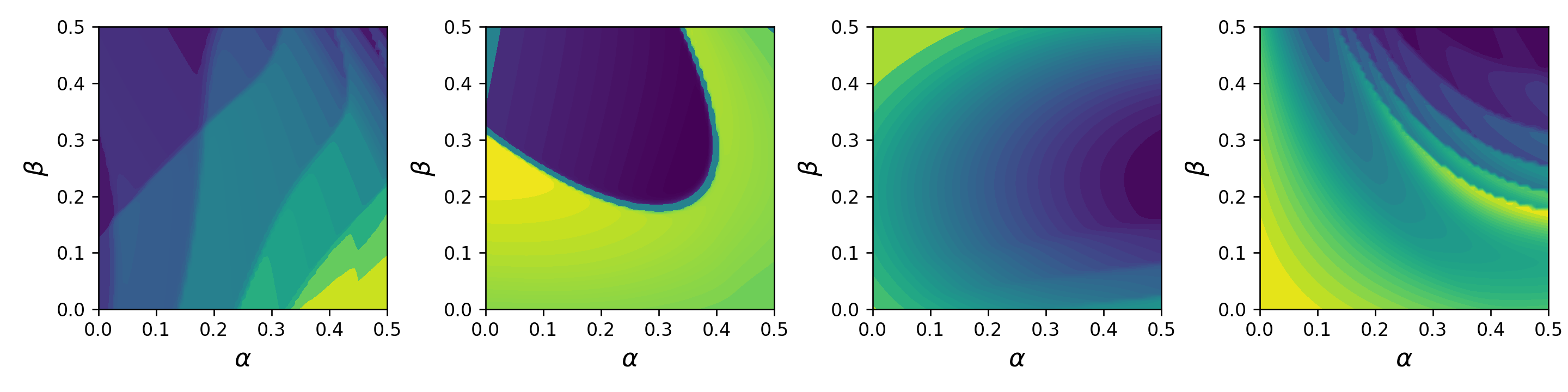}
	\caption{Random slices of the loss function for an E. coli cultivation. For each of the slices, three random parameter vectors are drawn from within the corresponding ranges, the loss function is evaluated on linear combinations of these parameters with coefficients $\alpha$ and $\beta$. For illustration purposes, levels are set such that each level contains roughly the same area.}
	\label{fig:slices}
\end{figure}

The optimisation is typically carried out by gradient or quasi-Newton
methods such as the BFGS algorithm~\citep[see, for example,][]{fletcher2000practical}.
Firstly, gradient of an ODE solution tends to take noticeably more
computational time than just obtaining the forward solution. Secondly,
the non-linear nature of the models tends to induce a highly complex
landscape of the loss function, Figure~\ref{fig:slices} illustrates this effect. This
might lead to a significant increase in the number of steps an optimiser
needs to perform for convergence. Moreover, the loss function seems to
have a significant number of local minima, which is typically addressed
by the multi-start algorithm~\citep{fletcher2000practical},
deterministic~\citep{lin2006deterministic} or nondeterministic~\citep{da2013comparison} global optimisation methods.
Multiple shooting also proved to reduce this problem in this context~\citep{peifer2007parameter}.
In this paper, we consider multi-start as our baseline as it offers a more direct
comparison with the proposed method.

All of the features of mechanistic models described above generally lead
to a high demand for computational resources when performing fitting.
Bayesian methods, like Monte-Carlo Markov chains, are fairly similar to
Maximum Likelihood fitting procedures in terms of computational
resources, although they might require substantially more computational
power. Additionally, please, notice that the mentioned algorithms
display a limited degree of parallelism: both gradient optimisation and
ODE solver are inherently sequential, albeit, methods like multiple
shooting partially alleviate the problem~\citep{gander2007analysis}

High usage of resources might limit online applications such as
monitoring, control and, especially, online design of experiments.
Currently, modern hardware seems to provide enough computational power
for such applications assuming a small number of unknown parameters and
good initial estimates \citep[for instance, see][]{kemmer2022nonlinear, krausch2022high},
nonetheless, more complex mechanistic models or a need for a more frequent response might prove difficult.

\section{Deep Learning for inference}

All of the fitting algorithms mentioned above are general-purpose
methods, they place weak assumptions on the loss function and work
practically with any reasonable model, treating the latter as a black
box that provides trajectories, values of the loss function and
gradients at arbitrary points etc.

\begin{figure}[t]
	\centering
	\begin{subfigure}{0.45\textwidth}
		\includegraphics[width=\textwidth]{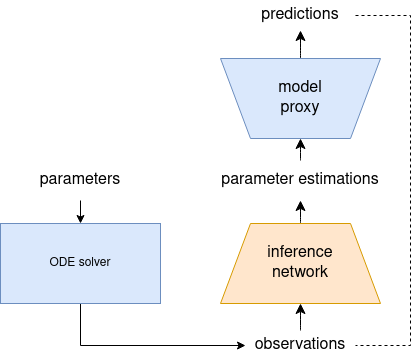}
	\end{subfigure}
	~
	\begin{subfigure}{0.45\textwidth}
		\includegraphics[width=\textwidth]{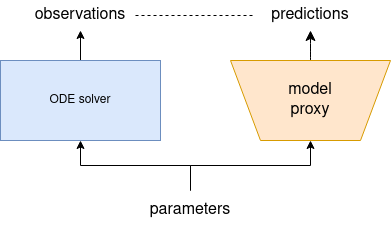}
	\end{subfigure}
	\caption{Schematic representation of the proposed method. Arrows
		represent flow of the data, ``inference network'' and ``model proxy''
		are represented by neural networks. Dashed lines indicate which
		quantities are used for the loss computation. (Left) Inference network
		training procedure, model proxy is frozen during training. (Right)
		``Model proxy'' training.}
	\label{fig:network}
\end{figure}

If one, however, considers a particular model, then the inference task,
Equation~\eqref{eq:ml}, can be viewed as a mapping from observations $x$ to the
optimal parameters $\theta^{*}(x)$:
\begin{align}
\theta^{\text{*}}\left( x \right) = \argmax_\theta\left[ -\log P\left( x \mid \theta \right) \right];
\end{align}
and, therefore, can be cast as a regression problem, with the loss
function:
\begin{align}
	\mathcal{L}(\psi) = -\sum_i \log P\left( x_i \Mid f(x_i \mid \psi) \right); \label{eq:loss}
\end{align}
where $f$ denotes the regressor parametrised by $\psi$. The regressor
$f$ can be represented by a neural network, which we call the
inference network. Please, notice that, unlike typical regression
problems, Equation~\ref{eq:loss} defines a loss function that closely resembles that of
an auto-encoder~\citep{kramer1991nonlinear}: the inference network provides parameter
estimates $\theta\left( x \right) = f\left( x\mid\psi \right)$, which
are then passed into the mechanistic model \emph{M} to make predictions
$\mu = M\left( f\left( x\  \mid \ \psi \right) \right)$, and, in a
sense, are ``decoded into'' predictions, and the latter are compared
against observations using the noise model $P_{n}\left( x\  \middle| \ \mu \right)$:
\begin{align}
- \log P\left(x \Mid \theta = f(x \mid \psi) \right) =
	- \log P_{n}\left(x \Mid \mu = M\left(f(x \mid \psi)\right)\right).
\end{align}

\subsection{Proxy network}
The major downside of the proposed algorithm is that networks
  typically require a large number of steps for training which greatly
  exceeds the number of training samples, and each step the algorithm
  requires evaluation of the mechanistic model and its gradient, which
  can not be precomputed as they depend on network predictions. In order
  to circumvent a need for the mechanistic model during NN training, we
  propose replacing the model with a so-called proxy network which is
  trained to approximate output of the mechanistic model. To train the
  proxy network we sample a large dataset of model parameters from the
  relevant part of the parameter space and corresponding observations,
  then train the network in a conventional manner. Once trained, the
  proxy network offers a fast approximation of the mechanistic model,
  which is used to guide the inference network (Figure~\ref{fig:network}). In our
  experiments, such a procedure results in decent inference networks,
  however, approximation errors of the proxy network prevent it from
  reaching precise solutions. Thus, we additionally retrain the
  inference network trained with a proxy network with the mechanistic
  model when it is not too costly.

\begin{figure}[t]
	\centering
	\includegraphics[width=4.77083in,height=3.57847in]{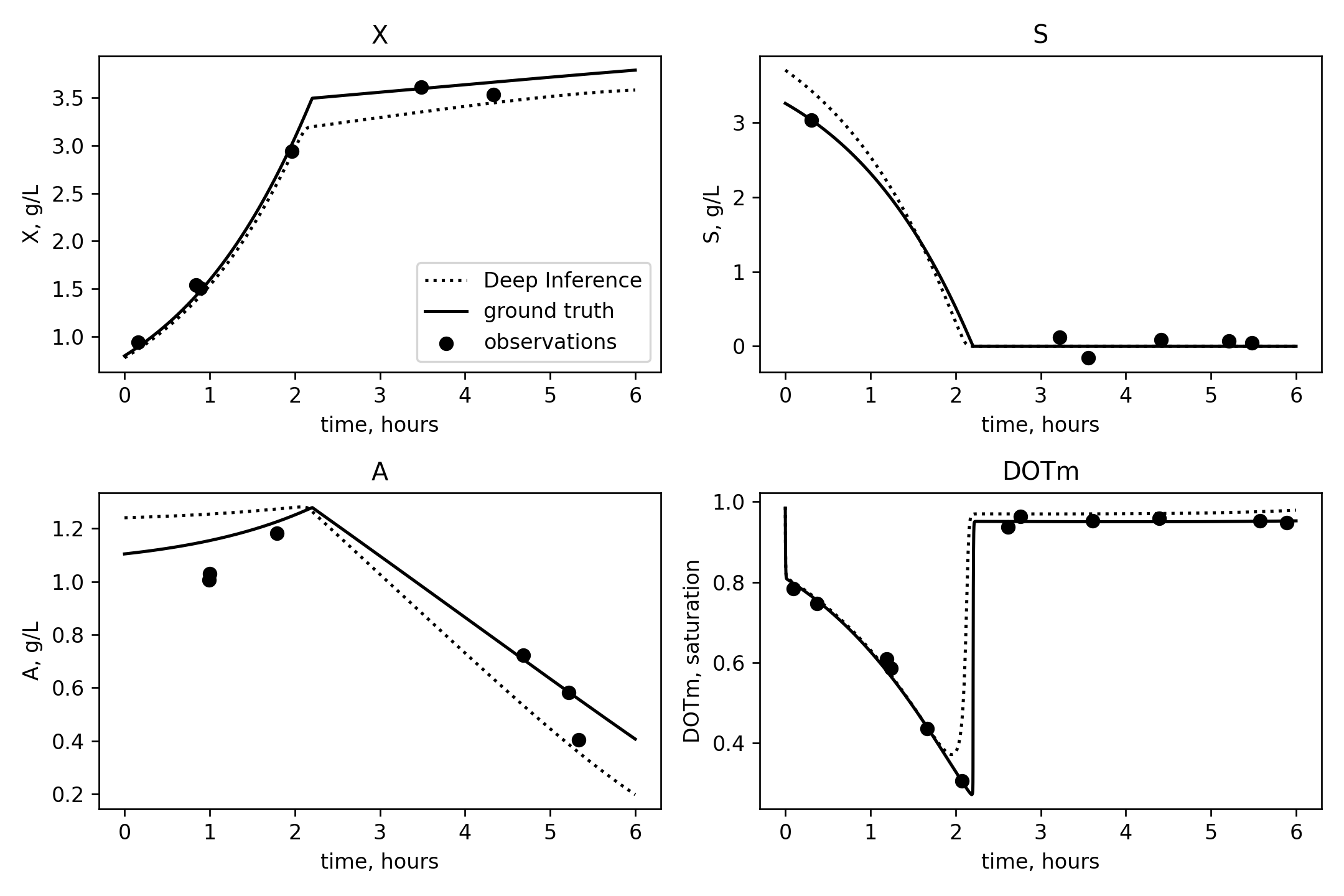}
	\caption{Deep Inference network evaluated on E. coli model by~\citet{anane2017modelling}.
		X, S and A denote biomass, substrate and acetate concentrations, DOTm – dissolved oxygen tension (as a faction of the saturation value).}
\end{figure}

\subsection{Numerical experiments}

To evaluate the performance of the proposed method, we simulate data
from two mechanistic models: Michaelis-Menten kinetics (MMK) and the E.
coli growth model by \citet{anane2017modelling}, both expressed as ODE systems.
For each model we sample parameters from a wide prior distribution,
obtain trajectories and from each trajectory we generate a small number
of observations for each of the observed channels, overall, 14
observations for MMK, 30 observations for the E. coli model.
Observations' timestamps are randomly distributed within the
corresponding range: 2 h for MMK, 6 hours for E. coli. For emulating
realistic conditions, channel observations are not aligned, i.e., only
one channel is measured per sample. For each of the models, we draw just
above a million training samples and around a thousand test ones.
Initial conditions are assumed to be unknown and inferred along with
models' parameters.

For the inference networks, we employ Deep Set architecture by~\citet{zaheer2017deep}
and use triplet encoding \citep{yalavarthi2022dcsf} for
processing asynchronous observations. To improve training, we apply an
invertible transformation that maps prior distribution of parameters and
initial conditions into the standard normal distribution, thus, bringing
all parameters to the same scale.

As a baseline we employ the BFGS optimisation algorithm~\citep{fletcher2000practical}.
We set a limit of 1024 iterations per sample
to reflect the real-world time constraints of an online application.
Additionally, we perform a multi-start procedure with a varying number
of initial guesses. To assess the improvement potential for the proposed
method, we also fine-tune predictions of the networks by running a
fitting algorithm starting from the networks' estimates.

\begin{table}[t]
	\setcellgapes{3pt}\makegapedcells
	\begin{tabular}[]{| l | c | c | c | c |}
		\hline
		\multirow{2}{*}{Method} & \multicolumn{2}{|c|}{MMK} & \multicolumn{2}{|c|}{E. coli} \\
		\cline{2-5}
		& $R^2$  & time & $R^2$ & time \\ \hline
		BFGS, 1 start & $0.928 \pm 0.010$ & \SI{50}{\milli\second} & $0.442 \pm 0.108$ & \SI{2.04}{\second} \\ \hline
		BFGS, 2 starts & $0.957 \pm 0.001$ & \SI{93}{\milli\second} & $0.930 \pm 0.016$ & \SI{3.59}{\second} \\ \hline
		BFGS, 4 starts & $0.958 \pm 0.001$ & \SI{180}{\milli\second} & $0.984 \pm 0.002$ & \SI{8.15}{\second} \\ \hline
		BFGS, 8 starts & $0.958 \pm 0.001$ & \SI{344}{\milli\second} & $0.989 \pm 0.001$ & \SI{16.5}{\second} \\ \hline
		Deep Inference & $0.949 \pm 0.001$ & \SI{10}{\micro\second} / \SI{37}{\micro\second}  & $0.945 \pm 0.007$ & \SI{18}{\micro\second} / \SI{260}{\micro\second} \\ \hline
		Deep Inference + MM & $0.954 \pm 0.001$ & \SI{10}{\micro\second} / \SI{37}{\micro\second} & - & -\\ \hline
		Deep Inference + BFGS & $0.958 \pm 0.001$ & \SI{41}{\micro\second} & $0.990 \pm 0.001$ & \SI{1.3}{\second}\\ \hline
	\end{tabular}
	\caption{Results of the numerical experiments. $R^2$ is
		computed based on squared errors normalised by variances of the noise in
		each individual channel. ``MM'' denotes fine-tuning with the
		corresponding mechanistic model, ``Deep Inference + BFGS'' denotes the
		fitting procedure (a single start) with initial guess produced by the
		network. Time measurements are per sample, GPU / CPU evaluations when applicable, CPU only otherwise.}
\end{table}

We evaluate performance of the algorithms on 1024 independently drawn
test samples. We also measure inference speed: BFGS‑based methods are
evaluated on a CPU (AMD Ryzen 5 5600X), all networks -- also on a GPU
(NVIDIA GeForce RTX 3070).

Results of our experiments are summarised in Table~1. Results show that
even for simple models, such as Michaelis--Menten kinetics, optimisation
algorithms are sometimes stuck in local minima (as shown by the gains of
the multi-start). The proposed method achieves a lower average loss than
a single fitting run and performance comparable to the multi-start
procedure. Fine-tuned predictions are on par with the best multi-start
results. Importantly, please, note the difference in the inference
speed: the network is faster by 3 orders of magnitude for the simplest
of the models, and by 6 orders for the E. coli model. Additionally, we
observe that the fitting procedure aided by the inference network tend
to converge faster.

\section*{Conclusion}
We present an alternative framework for quickly and reliably
estimating models' parameters given observations. Based on Deep
Learning techniques, it speeds up Maximum Likelihood estimations by
several orders of magnitude while preserving accuracy of
optimisation-based techniques. Additionally, our experiments indicate
that the proposed method does not suffer from the local minima
problem. The proposed method has a potential to significantly improve
online applications, such as monitoring, control of bioprocesses and
design of experiments. Moreover, fast inference enables the use of
much more complex and precise models, potentially leading to overall
improvement of bioprocess development.

\section*{Acknowledgements}

We gratefully acknowledge the financial support of the German Federal
Ministry of Education and Research (01DD20002A -- KIWI biolab).

\bibliography{DeepInference.bib}

\begin{thebibliography}{13}
\providecommand{\natexlab}[1]{#1}
\providecommand{\url}[1]{\texttt{#1}}
\expandafter\ifx\csname urlstyle\endcsname\relax
  \providecommand{\doi}[1]{doi: #1}\else
  \providecommand{\doi}{doi: \begingroup \urlstyle{rm}\Url}\fi

\bibitem[Anane et~al.(2017)Anane, Neubauer, Bournazou,
  et~al.]{anane2017modelling}
Emmanuel Anane, Peter Neubauer, M~Nicolas~Cruz Bournazou, et~al.
\newblock Modelling overflow metabolism in escherichia coli by acetate cycling.
\newblock \emph{Biochemical engineering journal}, 125:\penalty0 23--30, 2017.

\bibitem[Da~Ros et~al.(2013)Da~Ros, Colusso, Weschenfelder, de~Marsillac~Terra,
  De~Castilhos, Corazza, and Schwaab]{da2013comparison}
Simon{\'\i} Da~Ros, Gabriel Colusso, Thiago~A Weschenfelder, Lisiane
  de~Marsillac~Terra, Fernanda De~Castilhos, Marcos~L Corazza, and Marcio
  Schwaab.
\newblock A comparison among stochastic optimization algorithms for parameter
  estimation of biochemical kinetic models.
\newblock \emph{Applied Soft Computing}, 13\penalty0 (5):\penalty0 2205--2214,
  2013.

\bibitem[Fletcher(2000)]{fletcher2000practical}
Roger Fletcher.
\newblock \emph{Practical methods of optimization}.
\newblock John Wiley \& Sons, 2000.

\bibitem[Gander and Vandewalle(2007)]{gander2007analysis}
Martin~J Gander and Stefan Vandewalle.
\newblock Analysis of the parareal time-parallel time-integration method.
\newblock \emph{SIAM Journal on Scientific Computing}, 29\penalty0
  (2):\penalty0 556--578, 2007.

\bibitem[Kemmer et~al.(2022)Kemmer, Fischer, Wilms, Cai, Gro{\ss}, King,
  Neubauer, and Cruz-Bournazou]{kemmer2022nonlinear}
Annina Kemmer, Nico Fischer, Terrance Wilms, Linda Cai, Sebastian Gro{\ss},
  R~King, Peter Neubauer, and Mariano~Nicolas Cruz-Bournazou.
\newblock Nonlinear state estimation as tool for online monitoring and adaptive
  feed in high-throughput cultivations.
\newblock 2022.

\bibitem[Kramer(1991)]{kramer1991nonlinear}
Mark~A Kramer.
\newblock Nonlinear principal component analysis using autoassociative neural
  networks.
\newblock \emph{AIChE journal}, 37\penalty0 (2):\penalty0 233--243, 1991.

\bibitem[Krausch et~al.(2022)Krausch, Kim, Barz, Lucia, Gro{\ss}, Huber,
  Schiller, Neubauer, and Cruz~Bournazou]{krausch2022high}
Niels Krausch, Jong~Woo Kim, Tilman Barz, Sergio Lucia, Sebastian Gro{\ss},
  Matthias~C Huber, Stefan~M Schiller, Peter Neubauer, and Mariano~N
  Cruz~Bournazou.
\newblock High-throughput screening of optimal process conditions using model
  predictive control.
\newblock \emph{Biotechnology and Bioengineering}, 119\penalty0 (12):\penalty0
  3584--3595, 2022.

\bibitem[Lin et~al.(2001)Lin, Mathiszik, Xu, Enfors, and
  Neubauer]{lin2001determination}
HY~Lin, B~Mathiszik, B~Xu, S-O Enfors, and P~Neubauer.
\newblock Determination of the maximum specific uptake capacities for glucose
  and oxygen in glucose-limited fed-batch cultivations of escherichia coli.
\newblock \emph{Biotechnology and bioengineering}, 73\penalty0 (5):\penalty0
  347--357, 2001.

\bibitem[Lin and Stadtherr(2006)]{lin2006deterministic}
Youdong Lin and Mark~A Stadtherr.
\newblock Deterministic global optimization for parameter estimation of dynamic
  systems.
\newblock \emph{Industrial \& engineering chemistry research}, 45\penalty0
  (25):\penalty0 8438--8448, 2006.

\bibitem[Neubauer et~al.(2003)Neubauer, Lin, and
  Mathiszik]{neubauer2003metabolic}
Peter Neubauer, HY~Lin, and B~Mathiszik.
\newblock Metabolic load of recombinant protein production: inhibition of
  cellular capacities for glucose uptake and respiration after induction of a
  heterologous gene in escherichia coli.
\newblock \emph{Biotechnology and bioengineering}, 83\penalty0 (1):\penalty0
  53--64, 2003.

\bibitem[Peifer and Timmer(2007)]{peifer2007parameter}
Martin Peifer and Jens Timmer.
\newblock Parameter estimation in ordinary differential equations for
  biochemical processes using the method of multiple shooting.
\newblock \emph{IET systems biology}, 1\penalty0 (2):\penalty0 78--88, 2007.

\bibitem[Yalavarthi et~al.(2022)Yalavarthi, Burchert, and
  Schmidt-Thieme]{yalavarthi2022dcsf}
Vijaya~Krishna Yalavarthi, Johannes Burchert, and Lars Schmidt-Thieme.
\newblock Dcsf: Deep convolutional set functions for classification of
  asynchronous time series.
\newblock In \emph{2022 IEEE 9th International Conference on Data Science and
  Advanced Analytics (DSAA)}, pages 1--10. IEEE, 2022.

\bibitem[Zaheer et~al.(2017)Zaheer, Kottur, Ravanbakhsh, Poczos, Salakhutdinov,
  and Smola]{zaheer2017deep}
Manzil Zaheer, Satwik Kottur, Siamak Ravanbakhsh, Barnabas Poczos, Russ~R
  Salakhutdinov, and Alexander~J Smola.
\newblock Deep sets.
\newblock \emph{Advances in neural information processing systems}, 30, 2017.

\end{thebibliography}
\end{document}